# Gathering Statistics to Aspectually Classify Sentences with a Genetic Algorithm

Eric V. Siegel and Kathleen R. McKeown

Department of Computer Science
Columbia University
New York, NY 10027
evs@cs.columbia.edu, kathy@cs.columbia.edu

**Abstract.** This paper presents a method for large corpus analysis to semantically classify an entire clause. In particular, we use cooccurrence statistics among *similar clauses* to determine the aspectual class of an input clause. The process examines linguistic features of clauses that are relevant to aspectual classification. A genetic algorithm determines what combinations of linguistic features to use for this task.

## 1 Introduction

The aspectual class of a clause reflects the underlying temporal structure of the event(s) described by the phrase. For example, *"The boy held the bottle"* is a *non-telic event*, an event with no explicit temporal boundaries, while *"You will find us there"* describes a *telic event*, an event with a determined point of completion. Aspectual class is crucial to many natural language problems, including the determination of temporal constraints between events in a discourse [16, 18] as well as lexical choice and tense selection in machine translation [16, 12, 11, 4].

The number of ways in which clausal constituents interactively influence aspectual class is unknown. However, syntax alone is not sufficient, and the lexical head of multiple constituents (e.g., the verb phrase and the direct object) can be influential factors. Given the difficulty in identifying the precise factors determining the aspectual class of a clause, some researchers have used empirical analysis of corpora to develop probabilistic weights that aid in aspectual classification [12].

In this paper, we measure the ability for large corpus analysis to aid in semantically classifying an entire clause (e.g., a simple sentence). In particular, our approach uses cooccurrence statistics among *similar clauses* to determine the aspectual class of an input clause. Our approach integrates relevant linguistic knowledge with a statistical approach, using a genetic algorithm to control the acquisition of applicable statistics. In particular, the genetic algorithm creates a symbolic expression (i.e. *genetic programming*) that determines which linguistic features of each similar clause should be taken into account when determining aspect for a given clause; thus, the particular features that are used and their interaction can be determined flexibly. Experimental results show that our approach is viable and we identify extensions for further improving performance.

This work extends the statistical analysis of lexical cooccurrence to the statistical analysis of entire clauses. Lexical cooccurrences have been used for disambiguation problems [23, 14], and for the automatic identification of semantically related groups of words [19, 8]. Our work differs in that cooccurrence is counted among "similar" *clauses* as opposed to individual words, we use cooccurrence of clausal *features* (e.g., tense), and a symbolic expression created automatically by a genetic algorithm determines which features are of interest.

In the following sections, we first define the problem, identifying the aspectual classes we are interested in inferring. We then present our overall approach, showing how we compute clausal similarity, identifying the linguistic features that are used in classification, and showing how genetic programming is used for the task. Finally, we present experimental results and discuss conclusions and future work.

## 2  Aspectual Classification

Aspectual classification divides clauses along three dimensions. As shown in Table 1, a clause is classified as a *state* (e.g., *"Mark is happy"*) or an *event* (e.g., *"Renée ran down the street"*). Events are further divided according to *telicity*; that is, whether the event reaches a culminating point in time at which a new state is introduced. For example, *make* is usually **Telic**, since a new state is introduced – something is made. Events are also distinguished according to *atomiticity* (for more detail see Moens and Steedman [16]). The work described in this paper focuses on classifying events according to telicity.

Aspect must be considered when selecting a preposition for machine translation. *"John arrived late at work* **for** *several years"* describes a non-telic event, since John's repeated arrivals do not have one culminating point in time. As a result, its translation to French is, *"***Pendant** *des années Jean est arrivé en retard au travail."* However, *"John left the room* **for** *a few minutes"* describes a telic event. Here, the preposition *for* describes the duration of the state that results from the event. Therefore, the translation is, *"Jean a quitté la chambre* **pour** *quelques minutes."*[1] This demonstrates how the difference in telicity accounts for the different lexical choice in French.

The presence of certain arguments and adjuncts in a clause, as well as the tense of a clause, are constrained by and contribute to the aspectual class of the clause [5, 22, 20, 18, 11]. Examples of such constraints are listed in Table 2. Each entry in this table describes a syntactic *aspectual marker* and the constraints on the aspectual class of any clause that appears with that marker. For example, a telic event can be modified by an *in*-adverbial temporal adjunct, as in *"You will find us there* **in ten minutes***,"* but a non-telic event cannot, e.g., *"*\**The boy held the bottle* **in ten minutes***."* Known aspectual constraints are comprehensively summarized by Klavans [11].

Klavans and Chodorow pioneered the application of statistical corpus analysis to aspectual classification by placing verbs on a "stativity scale" according to the frequency with which they occur with aspectual markers [12]. This way, verbs are automatically ranked according to "how stative" they are.

---

[1] This example is from Moens and Steedman [16].

|  | EVENTS | | STATES |
|---|---|---|---|
|  | atomic | extended |  |
| Telic | CULMINATION | CULMINATED PROCESS |  |
| Non-Telic | POINT | PROCESS |  |

**Table 1.** Aspectual classes (drawn from Moens and Steedman [16]).

The aspectual class of a clause is determined by the verb, as well as other constituents of the clause. Therefore, one verb can describe more than one aspectual class, depending on its argumentative context. For example, the lexical head of the direct object can influence the clause's aspectual class: *"Sue played the piano"* signifies a non-telic event, while *"Sue played the sonata"* signifies a telic event [16].

## 3 Approach

Each occurrence of a linguistic marker from Table 2 constrains the aspectual class of the clause in which it appears. Our goal is to exploit the occurrences of these markers in a corpus to aid aspectual classification according to *telicity*.

Our approach is to examine the frequency with which linguistic markers occur among clauses that are *similar* to the input clause. *Similar* clauses are defined as those clauses that have the highest number of constituents in common, under the assumption that highly similar clauses will share the same aspectual class. This way, if many clauses that are similar to the input clause occur in the *perfect* tense, it is more likely that the input clause could occur in the *perfect* tense. This in turn denotes a high probability that the input clause is *telic*. We use a genetic algorithm to determine how to coordinate multiple aspectual indicators, as detailed below.

In this section we describe the measure of clausal similarity used for aspectual classification and the individual linguistic markers used. Next we describe how these components fit together to perform aspectual classification, and detail the use of a genetic algorithm. Finally, we specify the corpus and preprocessing used for this study.

### 3.1 Clausal Similarity

Two clauses that have syntactic constituents in common are likely to share the same aspectual class. For example, *"Peter drove the car"* and *"Sarah drove the bus"* both describe *non-telic* events. However, a prepositional phrase can also influence aspectual class: *"Peter drove the car to California"* describes a *telic* event. Therefore, multiple constituents must be considered when comparing clauses. In this study, we consider five clausal constituents when selecting similar clauses:

- adjunct preposition
- object determiner
- verb
- particle
- complement preposition

| If a clause can occur: | then it must be: |
|---|---|
| in *progressive* | **Extended Event** |
| in a *pseudo-cleft* | **Event** |
| with an *in-adverbial temporal adjunct* | **Telic Event** |
| in the *perfect* tense | **Telic Event** or **State** |

**Table 2.** Example aspectual markers.

These were empirically selected from a larger group of constituents based on their effectiveness in predicting aspectual class.

Since the selected group of similar clauses must be large enough to accurately measure the frequency of aspectual markers, we select 100 similar clauses from the corpus. However, it is unlikely, for any input clause, that there will be 100 clauses with all five constituents in common. Therefore, it is often necessary to relax this constraint. This is accomplished by considering one or more of the constraints to be a "wildcard," in which case it can be matched to any value. The system systematically generates all combinations of wildcards, from fewest to greatest, until a set of 100 similar clauses are found.

The extraction of similar sentences from a corpus is also used in example-based machine translation [17]; the translation of a sentence is guided by the translation of a similar sentence in a bilingual corpus.

### 3.2 Aspectual Indicators

Once 100 similar clauses have been extracted, the occurrences of aspectual markers can be examined to predict the aspectual class of the input clause. The markers used in this study are listed in Table 3, which shows the average frequency with which each marker appears among clauses that are similar to telic clauses, and among clauses that are similar to non-telic clauses. The SpecialPerfect indicator is used to count occurrences of the perfect tense that are *not* accompanied with the progressive tense. This is because any clause in the progressive can also appear in the perfect, even if it is non-telic, e.g., *"I have been running."* The allMatch indicator allows the detection of similar clauses with all five constituents in common with the input clause so that, for example, fully matching clauses can be weighted more heavily by the genetic algorithm.

### 3.3 Classifying with Indicator Frequency

Each of the five indicators listed in Table 3 may have predictive value, as indicated by their telic and non-telic frequencies. Our goal is to combine these indicators to increase the performance of aspectual classification. Previous efforts in corpus-based natural language classification have incorporated machine learning methods to coordinate multiple indicators, e.g., to classify adjectives according to markedness [9], and to perform accent restoration [23]. Klavans and Chodorow [12] describe why a weighted sum of multiple aspectual indicators could be advantageous.

A function that combines the frequencies of aspectual markers, as measured over the group of similar clauses, may be an over-simplified approach to this problem, since this would overlook the interaction of linguistic markers within each individual similar clause (for example, as captured with the SpecialPerfect indicator).

| Aspectual Marker: | Description: | Telic Frequency | Non-Telic Frequency |
|---|---|---|---|
| NotProgressive | Not progressive tense | 96.56 | 95.35 |
| SpecialPerfect | Perfect and not progressive | 6.17 | 4.66 |
| allMatch | Completely matches input clause | 1.57 | 1.18 |
| NotPresTense | Tense is not present | 20.15 | 18.78 |
| Past/Pres participle | Tense is past or present participle | 90.59 | 88.53 |

**Table 3.** Aspectual linguistic indicators.

Our approach is to use a genetic algorithm to create a symbolic expression consisting of logical and mathematical functions that operates on each similar clause to combine aspectual indicators. In this way, the interaction of indicators within each similar clause can be accommodated. Furthermore, since the genetic algorithm makes no *a priori* assumptions about the form of the solution, there is an increase in flexibility as compared to many other machine learning methods.

### 3.4 Combining Indicators With a Genetic Algorithm

Genetic algorithms [10, 7] work with a pool (*population*) of *individuals* (i.e. candidate solutions), stochastically performing *reproductive operations* on the individuals, guided by a measure of *fitness*, the domain objective we wish to optimize. The first step of the genetic algorithm is to create a random population of individuals. Next, new individuals are repeatedly generated. Two individuals (*parents*) are selected using the fitness measure. These individuals are randomly combined (*crossover*) to create a new individual. This new individual is then inserted into the population, replacing a randomly selected unfit individual. The output of the algorithm is the most fit individual derived after a fixed number of iterations.

In our work, an individual is an evolved program (i.e., genetic programming) [3, 13]. Each evolved program is a function tree composed of logical and mathematical operations and aspectual indicators. In particular, each internal node of an evolved tree is one of: *add, subtract, multiply, and, or, if-then-else*. The *if-then-else* operation takes three values (i.e. subtrees), and the other operations take two. Each leaf of an evolved function tree is one of five boolean terminals corresponding to the five aspectual indicators in Table 3. *Fitness* is defined as classification performance of the evolved program over the set of training examples.

For an input clause we wish to aspectually classify, the evolved function is evaluated once for each of the 100 similar clauses selected from the corpus. For each evaluation, the five aspectual terminals are assigned according to the corresponding features of the similar clause. Each evaluation results in an integer, since a sequence of arithmetic and logical functions are applied to binary values. The overall output, to be used for classification, is the sum of these 100 evaluations. Because the indicators in Table 3 produce larger values for telic clauses than for non-telic, the same can be expected for the overall output. Therefore, a simple threshold is selected to discriminate overall outputs such that classification performance is maximized.

Because the genetic algorithm is stochastic, each run may produce a different function tree. Runs of the genetic algorithm have a population size of 500, and end after 10,000 new individuals have been evaluated. The output of the algorithm is the best evolved program, along with the threshold established for classification over the training cases.

### 3.5 A Parsed Corpus

In our system, the automatic identification of individual constituents within a clause is necessary for both the extraction of similar clauses and the identification of the aspectual markers in Table 2. The English Slot Grammar (ESG) [15] has previously been used on large corpora to accumulate aspectual data [12]. ESG is particularly attractive for this task since its output describes a clause's deep roles, detecting, for example, the true subject and object of a passivized phrase.

Our experiments are performed across a 846,913 word corpus of 10 novels from which 75,289 clauses were parsed fully by ESG, with no self-diagnostic errors (ESG failed on some of these novels' complex sentences).

### 3.6 Manually Marking Supervised Data

To evaluate the performance of our system, we manually marked 574 clauses from the parsed corpus according to their aspectual class. These 574 were selected evenly across the 10 novels, and none have *be* as the main verb, since we are testing a distinction between non-statives. Of these, 75 were rejected because of parsing problems (verb or direct object incorrectly identified), and 94 rejected because they described states. This left 405 event clauses with which to evaluate classification performance.

We used linguistic tests that were selected for this task by Passonneau [18]. First, if a clause can be read in a *pseudo-cleft*, it is non-stative (i.e. an *event*), e.g., *"What its parents did was run off,"* and *"*What we did was know what is on the other side."* As an additional check for stativity, we also tested the clause with *"What happened was..."* Second, if a clause in the past progressive necessarily entails the past tense reading, the clause describes a non-telic event. For example, *"We were talking just like men"* (non-telic) entails that *"We talked just like men"*, but *"The woman was building a house"* (telic) does not necessarily entail that *"The woman built a house."*

## 4 Results

Two batches of runs of the genetic algorithm were performed. The first optimized for overall accuracy (six runs), and the second optimized for non-telic F-measure [21], which is described below (seven runs). Performance (*fitness*) was measured across half the manually marked clauses (*training cases*). Evolved programs were then evaluated over the other, "unseen" half of the marked data (*test cases*).

Simply classifying every clause as **Telic** achieves an accuracy of 68.0% over the 203 test cases, since 138 are telic. However, this approach classifies all non-telic clauses incorrectly, achieving a *non-telic recall* of 0.0%. This method serves as a baseline for comparison (*Baseline A*) since we are attempting to improve over an *uninformed* approach.[2]

---

[2] Similar baselines for comparison have been used for many classification problems [6], e.g., part-of-speech tagging [2, 1].

|  | Telic | | Non-Telic | | Overall |
|---|---|---|---|---|---|
|  | recall | precision | recall | precision | accuracy |
| Batch 1 | 92.1% | 72.7% | 26.4% | 61.2% | 71.1% |
| Baseline A | 100.0% | 68.0% | 0.0% | 100.0% | 68.0% |
| Baseline B | 73.6% | 68.0% | 26.4% | 32.0% | 58.5% |
| Batch 2 | 40.7% | 76.5% | 72.7% | 36.8% | 51.0% |
| Baseline C | 27.3% | 68.0% | 72.7% | 32.0% | 41.8% |

**Table 4.** A favorable tradeoff between telic and non-telic recall is achieved.

With the first batch of runs, our system achieved a favorable tradeoff between telic and non-telic recall, with no loss in overall accuracy, as compared to Baseline A. As shown in Table 4, our system correctly classified 26.4% of the non-telic clauses on average, compared to Baseline A's non-telic recall of 0.0%. Note that it is possible for an *uniformed* approach to achieve the same non-telic recall by arbitrarily classifying 26.4% of all clauses as **Non-Telic**, and the rest as **Telic**. However, this method (*Baseline B*) loses in comparison to our system both in overall accuracy (58.5%) and telic recall (73.6%).

A further tradeoff between telic and non-telic recall was achieved by the second batch of genetic algorithm runs. These runs optimized for the F-measure, a combined measure equally weighting non-telic recall and non-telic precision. As shown in Table 4, the average non-telic recall was 72.7%, and other measures compare favorably with a baseline system (*C*) that arbitrarily classifies 72.7% of all clauses as **Non-Telic**.

These results present an advantage for applications that weigh the identification of non-telic clauses more heavily than that of telic clauses. For example, a prepositional phrase denoting a duration with *for*, e.g., *"for a minute,"* describes the duration of a non-telic event, e.g., *"She ran for a minute,"* or the duration of the state that results from a telic event, e.g., *"She left the room for a minute."* That is, correctly identifying the use of *for* depends on identifying the aspectual class of the clause it modifies. A language understanding system that incorrectly classifies *"She ran for a minute"* as **Telic** will not detect that *"for a minute"* describes the duration of the *run* event. If this system, for example, summarizes the duration of events, it is particularly important to correctly classify non-telic events. In this case, our approach is advantageous.

The average accuracy attained by Batch 1 (optimizing for accuracy) was 71.1%, as shown in Table 4. Although this accuracy is higher than that of Baseline A, based on a binomial test this is not significant, which may be due to our small sample size.

## 5 Conclusions and Future Work

We have shown that the occurrences of linguistic markers among the *similar clauses* in large corpora reveals aspectual class. Namely, a favorable tradeoff between telic and non-telic recall was achieved while at the same time results provide evidence for a slight gain in overall accuracy. This is profitable for tasks that weigh the identification of non-telic clauses more heavily than telic clauses. A genetic algorithm was used to successfully create symbolic expressions describing how linguistic markers can be combined for this task.

Even more importantly, we have developed an expandable framework for integrating multiple knowledge sources (linguistic markers) for aspectual classification. This framework can be exploited by incorporating additional markers, such as the *in-adverbial* constraint from Table 2. Further, the definition of clausal similarity can be generalized to incorporate the semantic categories of lexical items (e.g., the verb and the direct object).

When aspectually classifying a clause, the current system uses the clause's constituents to identify a set of similar clauses. We plan to extend the system with rules that perform aspectual classification directly from clausal constituents. Finally, we plan to compare genetic programming to other machine learning methods for this task.

## Acknowledgments

This research is supported in part by the Columbia University Center for Advanced Technology in High Performance Computing and Communications in Health care (funded by the New York State Science and Technology Foundation), the Office of Naval Research under contract N00014-95-1-0745 and by the National Science Foundation under contract GER-90-24069.

Judith Klavans was very helpful regarding the formulation of our work, and Alexander Chaffee, Vasileios Hatzivassiloglou, and Dekai Wu regarding the evaluation of our results. For comments on an earlier draft of this paper, we would like to thank those people mentioned, as well as Dragomir Radev and Ruth Reeves. Finally, we would like to thank Andy Singleton for the use of his GPQuick software.